\documentclass[11pt,a4paper]{article}
\usepackage[hyperref]{emnlp2020}
\usepackage{times}
\usepackage{latexsym}

\usepackage{microtype}

\usepackage{wrapfig}
\usepackage{float}
\usepackage{microtype}
\usepackage{bm}
\usepackage{amsmath,amssymb,amsfonts}
\usepackage{enumerate}
\usepackage[linesnumbered,ruled]{algorithm2e}
\usepackage{subfigure}
\usepackage{graphicx}
\usepackage{multirow}
\usepackage{booktabs}
\usepackage{arydshln}

\usepackage{xcolor}

\usepackage{cases}

\usepackage{color}
\definecolor{bluei}{RGB}{0,150,255}
\definecolor{redi}{RGB}{183,15,18}
\definecolor{greeni}{RGB}{111,139,62}

\aclfinalcopy

\title{Mimic and Conquer: Heterogeneous Tree Structure Distillation \\for Syntactic NLP}

\author{
Hao Fei$^1$, Yafeng Ren$^2$ \and Donghong Ji$^1$\thanks{~~Corresponding author.} \\
1. Department of Key Laboratory of Aerospace Information Security and Trusted Computing,\\
Ministry of Education, School of Cyber Science and Engineering, Wuhan University, China \\ 
2. Guangdong University of Foreign Studies, China \\
\texttt{\{hao.fei,renyafeng,dhji\}@whu.edu.cn}\\
}

\date{}

\begin{document}
\maketitle
\begin{abstract}
Syntax has been shown useful for various NLP tasks, while existing work mostly encodes singleton syntactic tree using one hierarchical neural network.
In this paper, we investigate a simple and effective method, Knowledge Distillation, to integrate heterogeneous structure knowledge into a unified sequential LSTM encoder.
Experimental results on four typical syntax-dependent tasks show that our method outperforms tree encoders by effectively integrating rich heterogeneous structure syntax, meanwhile reducing error propagation, and also outperforms ensemble methods, in terms of both the efficiency and accuracy.
\end{abstract}

\section{Introduction}

Integrating syntactic information into neural networks has received increasing attention in natural language processing (NLP), which has been used for a wide range of end tasks, such as sentiment analysis (SA) \cite{nguyen2015phrasernn,TengZ17,LooksHHN17,zhang-zhang-2019-tree},
neural machine translation (NMT) \cite{cho-etal-2014-properties,garmash-monz-2015-bilingual,gu-etal-2018-top},
language modeling \cite{yazdani-henderson-2015-incremental,zhang-etal-2016-top,Zhou2017Generative},
semantic role labeling (SRL) \cite{marcheggiani-titov-2017-encoding,strubell-etal-2018-linguistically,Crossfei9165903},
natural language inference (NLI) \cite{tai-etal-2015-improved,liu-etal-2018-structured}
and text classification \cite{chen-etal-2015-sentence,zhang-etal-2018-graph}.
Despite the usefulness of structure knowledge, most existing models use only a single syntactic tree, such as a constituency or a dependency tree.

\begin{figure}[!t]
\centering \includegraphics[width=1.0\columnwidth]{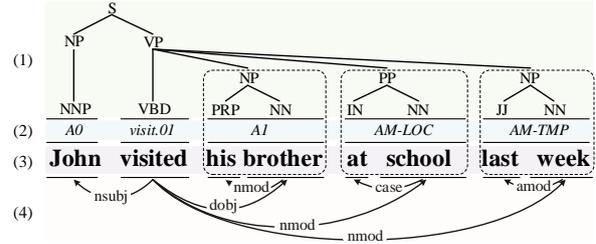}
\caption{
An example illustrating the mutual benefit of constituency and dependency tree structures.
(1) refers to the constituency tree structure, (2) indicates the semantic role labels, (3) refers to the example sentence, (4) represents the dependency tree structure.
}
\label{intro}
\end{figure}

Constituent and dependency representation for syntactic structure share underlying linguistic and computational characteristics, while differ also in various aspects.
For example, the former focuses on revealing the continuity of phrases, while the latter is more effective in representing the dependencies among elements.
By integrating the two representations from heterogeneous trees, the mutual benefit has been explored for joint parsing tasks \cite{collins-1997-three,charniak-johnson-2005-coarse,farkas-etal-2011-features,yoshikawa-etal-2017-ccg,zhou-zhao-2019-head}.
Intuitively, complementary advantages from heterogeneous trees can facilitate a range of NLP tasks, especially syntax-dependent ones such as SRL and NLI.
Taking the sentence of Figure \ref{intro} as example, where an example is shown from the SRL\footnote{We consider the span-based SRL, which aims to annotate the phrasal span of all semantic arguments.} task.
In this case, the dependency links can locate the relations between arguments and predicates more efficiently, while the constituency structure can aggregate the phrasal spans for arguments, and guide the global path to the predicate.
Integrating the features of two structures can better guide the model to focus on the most suitable phrasal granularity (as circled by the dotted box), and also ensure the route consistency between the semantic objective pairs.

In this paper, we investigate the \emph{Knowledge Distillation} (KD) method, which has been shown to be effective for knowledge ensembling \cite{HintonVD15,kim-rush-2016-sequence,FurlanelloLTIA18}, for heterogeneous structure integration.
Specifically, we employ a sequential LSTM as the student for distilling heterogeneous syntactic structures from various teacher tree encoders, such as GCN \cite{KipfW17} and TreeLSTM \cite{tai-etal-2015-improved}.
We consider output distillation, syntactic feature injection and semantic learning.
In addition, we introduce an alternative structure injection strategy to enhance the ability of heterogeneous syntactic representations within the shared sequential model.
The distilled structure-aware student model can make inference using sequential word inputs alone, reducing the error accumulation from external parsing tree annotations.

We conduct extensive experiments on a wide range of syntax-dependent tasks, including semantic role labeling, relation classification, natural language inference and sentiment classification.
Results show that the distilled student outperforms tree encoders, verifying the advantage of integrating heterogeneous structures.
The proposed method also outperforms existing ensemble methods and strong baseline systems, demonstrating its high effectiveness on structure information integration.

\section{Related Work}
 \subsection{Syntactic Structures for Text Modeling}

Previous work shows that integrating syntactic structure knowledge can improve the performance of NLP tasks \cite{socher2013recursive,cho-etal-2014-properties,nguyen2015phrasernn,LooksHHN17,liu-etal-2018-structured,zhang-zhang-2019-tree,FeiRJ20}.
Generally, these methods consider injecting either standalone constituency tree or dependency tree by tree encoders such as TreeLSTM
\cite{socher2013recursive,tai-etal-2015-improved} or GCN \cite{KipfW17}.
Based on the assumption that the dependency and constituency representation can be disentangled and coexist in one shared model, existing efforts are paid for joint constituent and dependency parsing, verifying the mutual benefit of these heterogeneous structures \cite{collins-1997-three,charniak-2000-maximum,charniak-johnson-2005-coarse,farkas-etal-2011-features,RenCK13,yoshikawa-etal-2017-ccg,strzyz-etal-2019-sequence,kato-matsubara-2019-ptb,zhou-zhao-2019-head}.
However, little attention is paid for facilitating the syntax-dependent tasks via integrating heterogeneous syntactic trees.
Although the integration from heterogeneous trees can be achieved via widely employed approaches, such as ensemble learning \cite{Wolpert92,ju10772} and multi-task training \cite{LiuQH16,ChenQLH18,FeiRJ20a}, they usually suffer from low-efficiency and high computational complexity.

 \subsection{Knowledge Distillation}

Our work is related to knowledge distillation techniques.
It has been shown that KD is very effective and scalable for knowledge ensembling \cite{HintonVD15,FurlanelloLTIA18}, and existing methods are divided into two categories:
1) \emph{output distillation}, which makes a teacher model output logits as a student model training objective \cite{kim-rush-2016-sequence,vyas-carpuat-2019-weakly,clark-etal-2019-bam},
2) \emph{feature distillation}, which allows a student to learn from a teacher's intermediate feature representations \cite{ZagoruykoK17,sun-etal-2019-patient}.
In this paper, we enhance the distillation of heterogeneous structures via both output and feature distillations by employing a sequential LSTM as the student.
Our work is also closely related to Kuncoro et al., \shortcite{kuncoro-etal-2019-scalable}, who distill syntactic structure knowledge to a student LSTM model.
The difference lies in that they focus on transferring tree knowledge from syntax-aware language model for achieving scalable unsupervised syntax induction, while we aim at integrating heterogeneous syntax for improving downstream tasks.
\section{Method}

\begin{figure}[!t]
\centering \includegraphics[width=0.98\columnwidth]{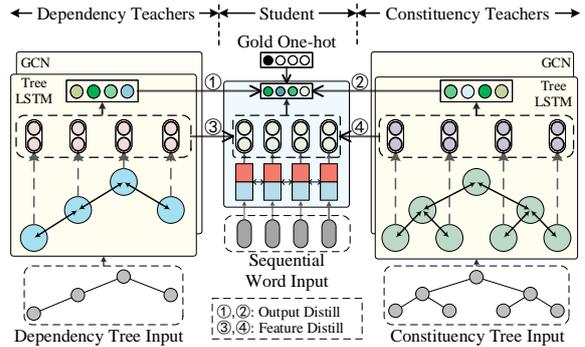}
\caption{
Overall framework of the proposed model.
}
\label{KD}
\end{figure}

As shown in Figure \ref{KD}, the overall architecture consists of a sequential LSTM \cite{HochreiterS97} student, and several tree teachers for dependency and constituency structures.

\subsection{Tree Encoder Teachers}

Different tree models can encode the same tree structure, resulting in different heterogeneous tree representations.
Following previous work \cite{tai2015improved,marcheggiani-titov-2017-encoding,zhang-zhang-2019-tree}, we consider encoding dependency trees by Child-Sum TreeLSTM and constituency trees by \emph{N}-ary TreeLSTM.
We also employ GCN to encode dependency and constituency structures separately.
We employ a bidirectional tree encoder to fully capture the structural information interaction.
Formally, we denote $\bm{X} = \{x_{1}, \cdots, x_{n}\}$ as an input sentence, $\bm{X}^\text{dep}= \{x_{1}^\text{dep}, \cdots, x_{n}^\text{dep}\}$ as the dependency tree and $\bm{X}^\text{con}= \{x_{1}^\text{con}, \cdots, x_{n}^\text{con}\}$ as the constituency tree.

\paragraph{Encoding dependency structure.}

We first use the standard Child-Sum TreeLSTM to encode the dependency structure, where each node $j$ in the tree takes as input the embedding vector $x_j$ corresponding to the head word.
The conventional bottom-up fashion is:
\begin{equation}
\begin{aligned} \label{dependency TreeLSTM}
\overline{h}_j &=  \sum_{k \in C(j)}   h_k \\
i_j &= \sigma(\bm{W}^{(i)} x_j^\text{dep} + \bm{U}^{(i)} \overline{h}_j + b^{(i)}) \\
f_{jk} &= \sigma(\bm{W}^{(f)} x_j^\text{dep} + \bm{U}^{(f)} \overline{h}_k + b^{(f)}) \\
o_j &= \sigma(\bm{W}^{(o)} x_j^\text{dep} + \bm{U}^{(o)} \overline{h}_j + b^{(o)}) \\
u_j &= \text{tanh}(\bm{W}^{(u)} x_j^\text{dep} + \bm{U}^{(u)} \overline{h}_j + b^{(u)}) \\
c_j &= i_j \odot u_j +  \sum_{k \in C(j)}  f_{jk} \odot c_k \\
h_j &= o_j \odot \text{tanh}(c_j)
\end{aligned}
\end{equation}
where $\bm{W}$, $\bm{U}$ and $b$ are parameters.
$C(j)$ refers to the set of child nodes of $j$.
$h_j, i_j, o_j$ and $c_j$ are the hidden state, input gate, output gate and memory cell of the node $j$, respectively.
$f_{jk}$ is a forget gate for each child $k$ of $j$.
$\sigma(\cdot)$ is an activation function and $\odot$ is element-wise multiplication.
Similarly, the top-down TreeLSTM has the same transition equations as the bottom-up TreeLSTM, except that the direction and the number of dependent nodes are different.
We concatenate the tree representations of two directions for each node: $h^{bi}_j = [ h^{\uparrow}_j ; h^{\downarrow}_j]$.

Compared with TreeLSTM, GCN is more computationally efficient in performing the tree propagation for each node in parallel with O(1) complexity.
Considering the constructed dependency graph
$\mathcal{G} = (\mathcal{V}, \mathcal{E})$,
where $\mathcal{V}$ are sets of nodes, and $\mathcal{E}$ are sets of bidirectional edges between heads and dependents, respectively.
GCN can be viewed as a hierarchical node encoder, representing the node $j$ at the $l$-th layer encoded as follows:
\begin{align}
\label{gcn gate} g_i^{l} &= \sigma(\bm{W}_i^{l} h_i^{l} + b_i^{l})  \\
\label{dependency GCN} h_j^l  &= \text{ReLU}( \begin{matrix} \sum_{i \in \mathcal{N}(i)}\end{matrix}  x_i^{l} \odot g_i^{l}  )
\end{align}
where $\mathcal{N}(i)$ are neighbors of the node $j$.
ReLU is a non-linear activation function.
For dependency encoding by TreeLSTM or GCN, we make use of all the node representations, $\bm{R}^{\text{dep}}=[r_1, \cdots, r_n]$, within the whole tree structure for next distillation.

\paragraph{Encoding constituency structure.}
We employ \emph{N}-ary TreeLSTM to encode constituent tree:
\begin{equation}
\begin{aligned} \label{constituency TreeLSTM}
i_j &= \sigma(\bm{W}^{(i)} x^\text{con}_j +  \sum_{q=1}^N \bm{U}_q^{(i)} h_{jq}  + b^{(i)}) \\
f_{jk} &= \sigma(\bm{W}^{(f)} x^\text{con}_j +  \sum_{q=1}^N \bm{U}_{kq}^{(f)} h_{jq} + b^{(f)}) \\
o_j &= \sigma(\bm{W}^{(o)} x^\text{con}_j +  \sum_{q=1}^N \bm{U}_q^{(o)} h_{jq}  + b^{(o)}) \\
u_j &= \text{tanh}(\bm{W}^{(u)} x^\text{con}_j +  \sum_{q=1}^N \bm{U}_q^{(u)} h_{jq}  + b^{(u)}) \\
c_j &= i_j \odot u_j +  \sum_{q=1}^N f_{jq}  \odot c_{jq} \\
h_j &= o_j \odot \text{tanh}(c_j)
\end{aligned}
\end{equation}
where $q$ is the index of the branch of $j$.
Slightly different from Child-Sum TreeLSTM, the separate parameter matrices for each child $k$ allow the model to learn more fine-grained and order-sensitive children information.
We also concatenate two directions from both bottom-up and top-down of each node as the final representation.

Similarly, GCN is also used to encode the constituent graph
$\mathcal{G} = (\mathcal{V}, \mathcal{E})$
via Eq. (\ref{gcn gate}) and (\ref{dependency GCN}).
Note that there are both words and constituent labels in the node set $\mathcal{V}$.
For constituency encoding by both TreeLSTM and GCN, we take the representations of terminal nodes in the structure as the corresponding word representations $\bm{R}^{\text{con}}=[r_1, \cdots, r_n]$.

\subsection{Heterogeneous Structure Distillation}
Sequential models have been proven effective on encoding syntactic tree information \cite{OrderedShenTSC18,kuncoro-etal-2019-scalable}.
We set the goal of KD as simultaneously distilling heterogeneous structures from tree encoder teachers into a LSTM student model.

We denote $\Gamma(\text{dep})$ = $\{\gamma(\text{\scriptsize{TreeLSTM}}), \gamma(\text{\scriptsize{GCN}})\}$ as the dependency teachers,
and $\Gamma(\text{con})$ = $\{\gamma(\text{\scriptsize{TreeLSTM}}), \gamma(\text{\scriptsize{GCN}})\}$ as the constituency teachers,
and $\Gamma(\text{all})$=$\Gamma(\text{dep}) \bigcup \Gamma(\text{con})$ as the overall teachers.
The objective of the student model can be decomposed into three terms: an output distillation target, a semantic target, and a syntactic target.

\paragraph{Output distillation.}

The output logits serve as \emph{soft target} providing richer supervision than the \emph{hard target} of one-hot gold label for the training \cite{HintonVD15}.
Given an input sentence $\bm{X}$ with the gold label $Y$ (one-hot), the output logits of teachers are $P^t_{\Gamma(\text{all})}$, and the output logits of the student is $P^s$.
The output distilling can be denoted as:
\begin{equation} \label{output}
    \mathcal{L}_{\text{output}} = \mathcal{H} ( [ \alpha Y + (1- \alpha) P^t_{\Gamma(\text{all})}  ] ,  P^s)
\end{equation}
where $\mathcal{H}(,)$ refers to the cross-entropy.
$\alpha$ is a coupling factor, which increases from 0 to 1 in training, namely \emph{teacher annealing} \cite{clark-etal-2019-bam}.

\paragraph{Syntactic tree feature distillation.}
In order to capture rich syntactic tree features, we consider allowing the student to directly learn from the teachers' feature hidden representation.
Specifically, we denote the hidden representation of the student LSTM as $\bm{R}^{\text{s}} = [r_1, \cdots, r_n]$, and we expect $\bm{R}^{\text{s}}$ to be able to predict the output of $\bm{R}^{\text{dep}}$ or $\bm{R}^{\text{con}}$ from syntax-aware teachers.
Thus the target is to optimize the following regression loss:
\begin{align}
\label{syn-A-dep} \mathcal{L}^{(A)}_{\text{dep}} &=  \frac{1}{2}    \sum_{j=1}^n  || f^{t}_{\Gamma(\text{dep})} (r_j^{\text{dep}}) \!-\! f^{s}(r_j^{\text{s}}) ||^2  \\
\label{syn-A-const} \mathcal{L}^{(A)}_{\text{con}} &= \frac{1}{2}   \sum_{j=1}^n  || f^{t}_{\Gamma(\text{con})} (r_j^{\text{con}}) \!-\! f^{s}(r_j^{\text{s}}) ||^2 \\
\label{syn-A} \mathcal{L}^{(A)}_{\text{syn}} &= \eta \mathcal{L}^{(A)}_{\text{dep}} + (1-\eta)  \mathcal{L}^{(A)}_{\text{con}}
\end{align}
where $\eta \in [0,1]$ is a factor for coordinating the dependency and constituency structure encoding, $f^{t}_{\Gamma(\text{dep})}()$, $f^{t}_{\Gamma(\text{con})}()$, $f^{s}()$ are the feedforward layers, respectively, for calculating the corresponding score vectors, and $j$ is the word index.

\paragraph{Semantic learning.}

We randomly mask a target input word $Q_j$ and let LSTM predict the word based on its hidden representation of prior words.
In consequence, we pose the following language modeling objective:
\begin{equation} \label{sem}
    \mathcal{L}_{\text{sem}} =  \sum_{j=1}^M  \mathcal{H}(Q_j, P^s_j|\bm{X}_{[1,\cdots,j-1]})
\end{equation}
by which LSTM can additionally improve the ability of semantic learning.

\subsection{Enhanced Structure Injection}
We consider further enhancing the trees injection, by encouraging the student to mimic the dependency and constituency tree induction of teachers.

\paragraph{Dependency injection.}
We force the student to predict the distributions of dependency arcs and labels based on the hidden representations and the representations of teachers.
\begin{equation}
\begin{aligned} \label{syn-B-dep}
   & \mathcal{L}^{(B)}_{\text{dep}} =    \sum_{j}^n \sum_{i}^n  \mathcal{H} ( P^t_{\Gamma(\text{dep})}(r_j|x_i),  P^s(r_j|x_i) ) \\
    &+   \sum_{j}^n \sum_{i}^n \sum_{k}^L  \mathcal{H} ( P^t_{\Gamma(\text{dep})}(l_k|r_j,x_i),  P^s(l_k|r_j,x_i) )
\end{aligned}
\end{equation}
where $P^t_{\Gamma(\text{dep})}(r_j|x_i)$ is the arc probability of the parent node $r_j$ for $x_i$ in the dependency teacher,
and $P^t_{\Gamma(\text{dep})}(l_k|r_j,x_i)$ is the probability of the label $l_k$ for the arc ($r_j, x_i$) in the teacher.

\paragraph{Constituency injection.}
Similarly, to enhance constituency injection, we mimic the distribution of each span $(i,j)$ with label $l$ in teachers.
Following Zhou et al. \shortcite{zhou-zhao-2019-head}, we adopt a feedforward layer as the span scorer:
\begin{equation} \label{syn-B-const-scorer}
    \text{Scr}(t) =   \sum_{(i,j)\in t} \sum_{k}   f(i,j,l)
\end{equation}
We use the CYK algorithm \cite{Cocke1970,Younger1975,Kasami1965} to search the highest score tree $T^*$ in teachers, and all possible trees $T$ in the student.
Then we optimize the following hinge loss between the structures in the student and teachers:
\begin{equation} \label{syn-B-const}
    \mathcal{L}^{(B)}_{\text{con}} =  \text{max}(0,  \mathop{\text{max}}\limits_{t \in T} (\text{Scr}(t) + \Delta(t, T^*)) - \text{Scr}(T^*) )
\end{equation}
where $\Delta$ is the hamming distance.
The above syntax loss in Eq. (\ref{syn-B-dep}) and (\ref{syn-B-const}) can substitute the ones in Eq. (\ref{syn-A-dep}) and (\ref{syn-A-const}), respectively.
The overall objective of the structure injection is:
\begin{equation} \label{syn-B}
    \mathcal{L}^{(B)}_{\text{syn}} = \eta \mathcal{L}^{(B)}_{\text{dep}}  + (1-\eta) \mathcal{L}^{(B)}_{\text{con}}
\end{equation}

\paragraph{Regularization.}
Based on the independent assumption, the syntax feature distillations target learning diversified private representations for heterogeneous structures as much as possible.
In practice, there should be a latent shared structure in the parameter space, while the separate distillations will squeeze such shared feature, weakening the expression of the learnt representations.
To avoid this, we additionally impose a regularization on Eq. (\ref{syn-A-dep}), (\ref{syn-A-const}), (\ref{syn-B-dep}) and (\ref{syn-B-const}): 
\begin{equation}
\mathcal{L}_{\text{reg}}=\frac{\zeta}{2}
 ||\Theta||^{2} \,,
\end{equation}
where $\Theta$ is the overall parameter in the student.

\subsection{Training}

\textbf{Algorithm} \ref{process} gives the overall structure distillation process.
At early training stage (line 2-19), semantic learning (Eq. \ref{sem}) and output distillation (Eq. \ref{output}) are first executed by each teacher.
As we have multiple teachers for one student on each task, for syntactic tree structure distillation, we sequentially distill one teacher at one time.
We take turn with a turning gap $G_2$ processing the dependency or constituency injection from a tree teacher (line 13-17), to keep the training stable.
After a certain number of training iterations $G_1$, we optimize the overall loss (line 20):
\begin{equation} \label{all}
    \mathcal{L}_{\text{all}} =  \mathcal{L}_{\text{output}} + \lambda_1 \mathcal{L}_{\text{syn}} + \lambda_2 \mathcal{L}_{\text{sem}}
\end{equation}
where $\lambda_1$ and $\lambda_2$ are coefficients, which regulate the corresponding objectives.
$\mathcal{L}_{\text{syn}}$ can be either $\mathcal{L}_{\text{syn}}^{(A)}$ (Eq. \ref{syn-A}) or $\mathcal{L}_{\text{syn}}^{(B)}$ (Eq. \ref{syn-B}), that is, the syntax sources are simultaneously from two tree encoders (dependency and constituency) at one time.
During inference, the well-trained student can make prediction alone with only word sequential input.

\begin{algorithm}[!t]
\caption{Distill heterogeneous trees.}
\label{process}
\KwIn{{ Training set: ${(\bm{X},\bm{X}^\text{dep},\bm{X}^\text{con},Y)}$; Total iteration $T$; Syntax turn gaps $G_1$, $G_2$; Syntax flag $F$=Ture.}}
\KwOut{Student model. }
\While{$t < T$}{
    \eIf{$t \le G_1 $}{
        \If{$t \% G_2$ == 0}{
            $F \gets !F$ \;
        }
        $P^s \gets$ Student($\bm{X}$) \;
        {opt} $\mathcal{L_{\text{sem}}}$ in Eq. (\ref{sem}) \;
        \For{$\gamma(\text{\scriptsize{model}}) \in \Gamma(\text{all})$}{
            $P^t_{\Gamma(\text{dep})}, r_j^{\text{dep}} \gets \gamma(\text{\scriptsize{model}})(\bm{X}^\text{dep})$\;
            $P^t_{\Gamma(\text{con})}, r_j^{\text{con}} \gets \gamma(\text{\scriptsize{model}})(\bm{X}^\text{con})$\;
            $P^t_{\Gamma(\text{all})} = P^t_{\Gamma(\text{dep})} \bigcup P^t_{\Gamma(\text{con})}$ \;
            {opt} $\mathcal{L_{\text{output}}}$ in Eq. (\ref{output}) \;
            \eIf {$F$}{
                {opt} $\mathcal{L_{\text{dep}}}$ in Eq. (\ref{syn-A-dep}) or (\ref{syn-B-dep}) ;  
                // \emph{dependency learning}
            }{
                {opt} $\mathcal{L_{\text{con}}}$ in Eq. (\ref{syn-A-const}) or (\ref{syn-B-const}) ;  
                 // \emph{constituency learning}
            }
        }
    }{
        {opt} $\mathcal{L_{\text{all}}}$ in Eq. (\ref{all}) \; 
    }
}
\end{algorithm}

\section{Experiments}

\subsection{Experimental Setups}
\paragraph{Hyperparameters.}
We use a 3-layer BiLSTM as our student, and a 2-layer architecture for all tree teachers.
The default word embeddings are initialized randomly, and the dimension of word embeddings is set as 300.
The hidden size is set to 350 in the student LSTM, and 300 in the teacher models, respectively.
We adopt the Adam optimizer with an initial learning rate of 1e-5.
We use the mini-batch of 32 within total 10k ($T$) iterations with early stopping, and apply 0.4 dropout ratio for all embeddings.
We set the coefficients $\lambda_1$, $\lambda_2$, $\zeta$ and $\eta$ as 0.6, 0.2, 0.2 and 0.5, respectively.
The training iteration thresholds $G_1$ and $G_2$ are set as 300 and 128, respectively.
These values achieve the best performance in the development experiments.

\paragraph{Baselines systems.}
We compare the following baselines.
1) Sequential encoders: LSTM, attention-based LSTM (A{\scriptsize TT}LSTM) and Transformer \cite{vaswani2017attention}, sentence-state LSTM (S-LSTM) \cite{zhang-etal-2018-sentence};
2) Tree encoders introduced in $\S$2;
3) Ensemble models: ensembling learning (EnSem) \cite{Wolpert92,ju10772}, multi-task method (MTL) \cite{LiuQH16,ChenQLH18}, adversarial training (AdvT) \cite{liu-etal-2017-adversarial} and tree communication model (TCM) \cite{zhang-zhang-2019-tree}.
For EnSem, we only concatenate the output representations of tree encodes.
For MTL, we use an underlying shared LSTM for parameter sharing for tree encodes.
For AdvT, we adopt the shared-private architecture \cite{liu-etal-2017-adversarial} based on MTL.
Following Zhang et al. \shortcite{zhang-zhang-2019-tree}, for TCM, we initialize GCN for TreeLSTM, to encode dependency and constituency trees respectively, and finally concatenate the output representations.
Note that all the models in Tree Ensemble group take total four Tree teachers as in our distillation teachers' meta-encoders.
4) Other baselines: ESIM \cite{chen-etal-2017-enhanced}, local-global pattern based self-attention networks (LG-SANs) \cite{xu-etal-2019-leveraging} and BERT.

\paragraph{Tasks and evaluation.}
The experiments are conducted on four representative syntax-dependent tasks:
1) \texttt{Rel}, relation classification on Semeval10 \cite{hendrickx2009semeval};
2) \texttt{NLI}, sentence pair classification on the Stanford NLI \cite{bowman-etal-2015-large};
3) \texttt{SST}, binary sentiment classification task on the Stanford Sentiment Treebank \cite{socher2013recursive},
4) \texttt{SRL}, semantic role labeling on the CoNLL2012 OntoNotes \cite{pradhan-etal-2013-towards}.
For \texttt{NLI}, we make element-wise production, subtraction, addition and concatenation of two separate sentence representations as a whole.
We mainly adopt F1 score to evaluate the performance of different models.
The data splitting follows previous work.

\paragraph{Trees annotations and resources.}
The OntoNotes data offers the annotations of the dependency and constituency structure.
For the rest datasets, we parse sentences via the state-of-the-art BiAffine dependency parser \cite{BiaffineDozatM17}, and the Self-Attentive constituency parser \cite{kitaev-klein-2018-constituency}.
The parsers are trained on PTB\footnote{\url{https://catalog.ldc.upenn.edu/LDC99T42}.}.
The dependency parser has a 93.4\% LAS, and the constituency parser has 92.6\% F1 score.
Besides, we evaluate different contextualized word representations, such as ELMo\footnote{\url{https://allennlp.org/elmo}} and BERT\footnote{\url{https://github.com/google-research/bert}}.

\begin{table}[!t]
\begin{center}
\resizebox{1.0\columnwidth}{!}{
  \begin{tabular}{lllll}
\hline
 &    \bf Rel &  \bf NLI  &  \bf SST  & \bf  SRL \\
\hline
\multicolumn{5}{l}{ $\bullet$ \textbf{Sequential Encoder }}\\
\quad LSTM &  80.5 &  79.6 &  82.3 &  76.6 \\
\quad A{\scriptsize TT}{LSTM} &   82.3 &  81.5 &  84.2 &  78.2 \\
\quad Transformer &   84.7 &  84.2 &  85.0 &  80.5 \\
\quad S-LSTM &   85.0 &  84.8 &  86.2 &  82.0 \\
\hline
\multicolumn{5}{l}{$\bullet$ \textbf{Standalone Tree Model  }}\\
\quad TreeLSTM+dep. &   85.2 &  86.0 &  86.4 &  82.5 \\
\quad GCN+dep. &  85.9 &  85.8 &  86.1 &  83.3 \\
\quad TreeLSTM+con. &   85.0 &  86.8 &  87.6 &  82.2 \\
\quad GCN+con. &  84.8 &  86.3 &  86.8 &  81.8 \\
\hdashline
\quad \emph{Avg.}  &  85.3 &  86.2 &  86.5 &  82.4 \\
\hline\hline
\multicolumn{5}{l}{$\bullet$ \textbf{Tree Ensemble  }}\\
\quad EnSem &   85.5 &  87.0 &  86.0 &  81.4  \\
\quad MTL  & 84.9 &   88.3 &  87.2 &  83.7  \\
\quad AdvT &  86.4 &  87.6 &  85.2 &  82.1  \\
\quad TCM  & 85.7 &   88.8 &  88.4 &  83.0  \\
\hdashline
\quad \emph{Avg.} &  85.9 &   88.1 &  86.7 &  82.3 \\
\hline\hline
\multicolumn{5}{l}{$\bullet$ \textbf{Distilled Student }}\\
\quad \texttt{Best}   & \textbf{89.2$^{*}$}   & \textbf{90.8$^{*}$} &
\textbf{91.6$^{*}$} &   \textbf{85.5$^{*}$} \\
\hline\hline
\multicolumn{5}{l}{$\bullet$ \textbf{Others }}\\
\quad  ESIM  &  -  & 88.9 & - &  - \\
\quad  LG-SANs  &  85.6 &  86.5 &  87.3 &  81.2 \\
\quad  BERT  & \underline{91.3}  & \underline{92.1} & \underline{94.4} &  \underline{86.0} \\
\hline
\end{tabular}
}
\end{center}
\caption{
Main results on various end tasks.
$*$ indicates $p\le$0.05.
}
\label{heterogeneous tree for end task}
\end{table}

\subsection{Main Results}
Experimental results of different models are shown in Table \ref{heterogeneous tree for end task}, where several observations can be found.
First, tree models encoded with syntactic knowledge can facilitate syntax-dependent tasks, outperforming sequential models by a substantial margin.
Second, different tree encoders integrated with varying syntactic tree structures can make different contributions to the tasks.
For example, GCN with dependency structure gives the best result for \texttt{Rel}, while TreeLSTM with constituency tree achieves the best performance for \texttt{SST}.
Third, when integrating heterogeneous tree structures by tree ensemble methods, a competitive performance can be obtained, showing the importance of integrating heterogeneous tree information.
Finally, our distilled student model significantly outperforms all the baseline systems\footnote{Note that a direct comparison with BERT is unfair, because a large number of pre-trained parameters can bring overwhelming improvement.}, demonstrating its high effectiveness on the integration of heterogeneous structure information.

\begin{table}[!t]
\begin{center}
\resizebox{0.89\columnwidth}{!}{
  \begin{tabular}{lllll}
\hline
 &    \bf Rel &  \bf NLI  &  \bf SST  & \bf  SRL \\
\hline
\multicolumn{5}{l}{$\bullet$  Syntax Injection Strategy}\\
\quad  +{$\mathcal{L}_{syn}^{(A)}$}   & 88.6  & 90.2  & 91.0 &   85.0 \\
\quad  +{$\mathcal{L}_{syn}^{(B)}$}   & \underline{89.2}   & \underline{90.8} &
\underline{91.6} &   \underline{85.5} \\
\hline
\multicolumn{5}{l}{$\bullet$  Distilling Objective  (with $\mathcal{L}_{syn}^{(B)}$) }\\
\quad  w/o {$\mathcal{L}_{sem}$}   & 87.9  & 89.2 & 89.7 &  84.8 \\
\quad  w/o {$\mathcal{L}_{syn}$}   & 86.8 &   88.7 &  88.9 &  83.7 \\
\quad  w/o $\mathcal{L}_{{reg}}$   & 88.1 &   89.3 &  89.9 &  84.7 \\
\quad  w/o \scriptsize{Tea.Anl.} &   88.2 &  89.1 &  90.4 &  84.5 \\
\hline
\multicolumn{5}{l}{$\bullet$  Contextualized Semantics  (with $\mathcal{L}_{syn}^{(B)}$) }\\
\quad  +ELMo &   {90.6} &  {91.6} &  {92.4} &  {85.1} \\
\quad  +BERT  & \bf 92.2  & \bf 93.0 & \textbf{95.1} &  \bf 86.8 \\
\hline
\end{tabular}
}
\end{center}
  \caption{
  Ablation results on distilled student.
  `Tea.Anl.' refers to teacher annealing.
  In `Semantics', we replace semantic learning $\mathcal{L}_{sem}$ with pre-trained contextualized word representations.
  }
  \label{Ablation}
\end{table}

\paragraph{Ablation results.}
We ablate each part of our distilling method in Table \ref{Ablation}.
First, we find that the enhanced structure injection strategy ($\mathcal{L}_{syn}^{(B)}$) can help to achieve the best results for all the tasks, compared with the latent syntax feature mimic ($\mathcal{L}_{syn}^{(A)}$).
By ablating each distilling objective, we learn that the syntax tree distillation ({$\mathcal{L}_{syn}$}) is the kernel of our knowledge distillation for these syntax-dependent tasks, compared with semantic feature learning ({$\mathcal{L}_{sem}$}).
Besides, both the introduced teacher annealing factor $\alpha$ and regularization $\mathcal{L}_{{reg}}$ can benefit the task performance.
Finally, we explore recent contextualized word representations, including ELMo \cite{PetersNIGCLZ18} and BERT \cite{devlin-etal-2019-bert}.
Surprisingly, our distilled student model receives a substantial performance improvements in all tasks.
However, when removing the proposed syntax distillation from BERT, the performance drops, as shown in Table \ref{heterogeneous tree for end task} (the vanilla BERT).

\begin{table}[!t]
\begin{center}
\resizebox{0.96\columnwidth}{!}{
  \begin{tabular}{lcc}
\hline
 &   \bf Constituency &   \bf Dependency\\
\hline
 TreeLSTM+dep. &   28.31 &   73.92 \\
 GCN+dep. &   19.11 &   \textbf{76.32} \\
\hdashline
 TreeLSTM+con. &   \textbf{68.65} &   30.27 \\
 GCN+con. &   66.30 &   23.85 \\
\hdashline
 Student-Full &   \underline{62.61} &   \underline{70.34} \\
\quad w/o $\mathcal{L}_{{reg}}$ &   {53.20} &   {64.08} \\
\hline
\end{tabular}
}
\end{center}
\caption{
  Probing the upper-bound of constituent and dependent syntactic structure.
  }
  \label{Probing task}
\end{table}

\subsection{Heterogeneous Tree Structure}

\paragraph{Upper-bound of heterogeneous structures.}
We explore to what extent the distilled student can manage to capture heterogeneous tree structure information.
Following previous work \cite{Probing2018}, we consider employing two syntactic probing tasks, including
1) \textbf{Constituent labeling}, which assigns a non-terminal label for text spans within the phrase-structure (e.g., \emph{Verb}, \emph{Noun}, etc.),
and 2) \textbf{Dependency labeling}, which predicts the relationship (edge) between two tokens (e.g., \emph{subject-object} etc.).
We take the last-layer output representation as the probing objective.
We compare the student model with four teacher tree encoders, separately, based on the SRL task.
As shown in Table \ref{Probing task}, the student LSTM gives slightly lower score than one of the best tree models (i.e., \emph{GCN+dep.} for dependency labeling, \emph{TreeLSTM+con.} for constituency labeling), showing the effectiveness on capturing syntax.
Besides, we can find that the regularization $\mathcal{L}_{{reg}}$ plays a key role in improving the expression capability of the learnt representation.

\begin{figure}[!t]
\centering \includegraphics[width=0.9\columnwidth]{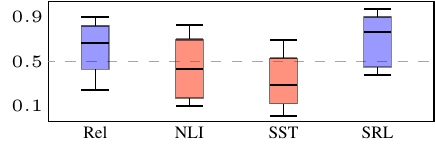}
\caption{
Heterogeneous syntax distribution.
The predominance of dependency syntax is above 0.5, otherwise for constituency.
}
\label{syntax-distribution}
\end{figure}

\begin{figure}[!t]
\centering \includegraphics[width=1.0\columnwidth]{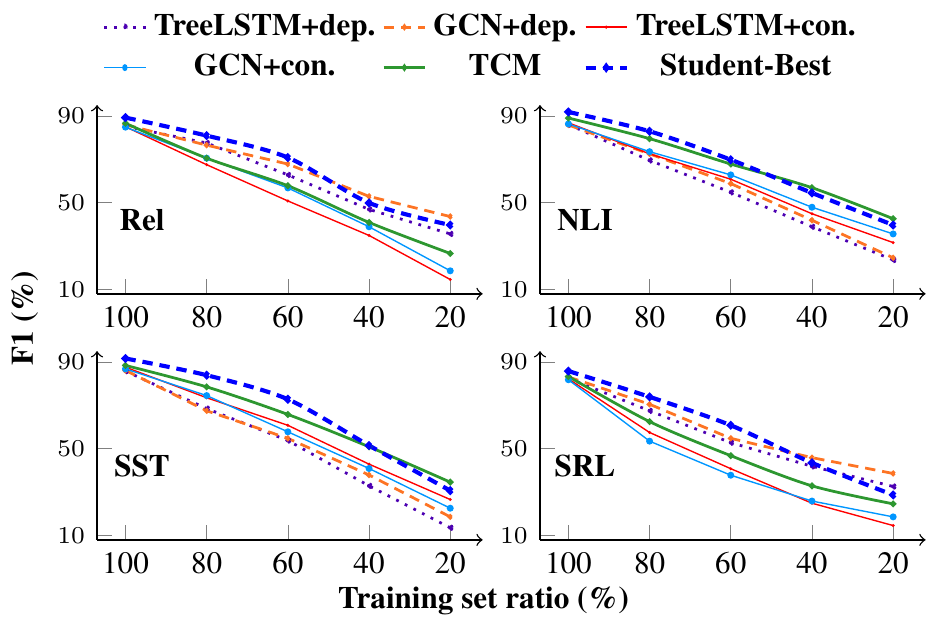}
\caption{
Results under varying ratio of train set.
}
\label{Generalization}
\end{figure}

\paragraph{Distributions of heterogeneous syntax in different tasks.}

We also compare the distributions of dependency and constituency structures in different tasks after fine-tuning.
Technically, based on each example in the test set, the performance drops when the student LSTM is trained only under either standalone dependency or constituency injection (TreeLSTM or GCN), respectively, by controlling $\eta$=0 or 1.
Intuitively, the more the results drop, the more the model benefits from the corresponding syntax.
For each task, we collect the sensitivity values and linearly normalize them into [0,1] for all examples, and make statistics.
As plotted in Figure \ref{syntax-distribution}, the distributions of dependency and constituency syntax vary among tasks, verifying that different tasks depend on distinct types of structural knowledge, while integrating them altogether can give the best effects.
For example, \texttt{TreeDepth}, dependency structures support \texttt{Rel} and \texttt{SRL}, while \texttt{NLI} and \texttt{SST} benefit from constituency the most.

\subsection{Robustness Analysis}

\paragraph{Generalization ability to training data.}

Figure \ref{Generalization} shows the performance of different models on varying ratio of the full training dataset.
We can find that the performance decreases with the reduction of the training data for all methods, while our distilled student achieves better results, compared with most of the baselines.
The underlying reasons are two-fold.
First, the heterogeneous syntactic features can provide strong representations for supporting better predictions.
Second, the distilled student takes only sequential inputs, avoiding the noise from parsed inputs to some extent.

Also we see that \emph{TreeLSTM/GCN+dep.} can counteract the data reduction ($\le$40\%) on \texttt{Rel} and \texttt{SRL} tasks, showning that they rely more on dependency structures, while \texttt{NLI} and \texttt{SST} depend on constituency structures.
In addition, the student starts underperforming than the best one on the small data ($\le$40\%).
Without explicit tree annotations, the contribution of heterogeneous syntax can be deteriorated.
But it still remains robust on shortage of training data than most of the baselines, due to its noise resistant.

\paragraph{Reducing error accumulation of tree annotation.}
We investigate the effects on reducing noises from tree annotation.
We compare the performance under different sources.
Table \ref{Parsing Error Reduction} shows the results on \texttt{SRL}.
With only word inputs, our model still outperforms the baselines which take the gold syntax annotation.
This partially shows that without parsed tree annotation, the student model can avoid noise and error propagation.
When we add gold annotation as additional signal, the performance can be further improved.

\begin{table}[!t]
\begin{center}
\resizebox{1.0\columnwidth}{!}{
\begin{tabular}{lccc}
\hline
\multicolumn{1}{l}{\bf System} &    \bf Auto-Syn &    \bf Gold-Syn &  w/o \bf Syn \\
\hline
\quad TreeLSTM+dep.  &   80.6 & 82.5 &  - \\
\quad GCN+dep. &  81.1 &  83.3  & -\\
\quad TreeLSTM+con.&  79.6 &  82.2 & -  \\
\quad GCN+con. &  79.8 &  81.8 &  - \\
 \hdashline
\quad EnSem  &   80.5 &  81.4  & - \\
\quad MTL &  81.2 &  83.7  & -  \\
\quad AdvT &  81.0 &  82.1 &  - \\
\quad TCM &   82.4 &  83.0  &  - \\
 \hdashline
\quad Student-Full &  - & \bf 86.2$^\dag$ &   \bf{85.5} \\
\hline
\end{tabular}
}
\end{center}
\caption{
Performance of different systems with automatically-parsed/gold syntax, and without syntax annotations.
$\dag$ indicates that we concatenate additional gold syntactic label with other input features.
}
\label{Parsing Error Reduction}
\end{table}

\begin{figure}[!t]
\centering \includegraphics[width=0.92\columnwidth]{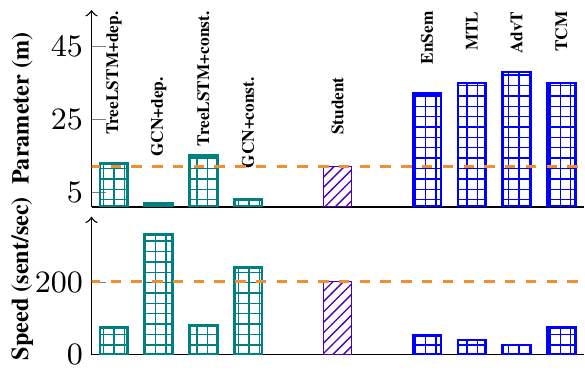}
\caption{
Comparisons on parameter scale and decoding speed.
}
\label{efficiency}
\end{figure}

\paragraph{Efficiency study.}
As shown in Figure \ref{efficiency}, the student model has fewer parameters, while keeping faster decoding speed, compared with other ensemble models.
Our sequential model is about 3 times smaller than AdvT, but nearly 4 times faster than the tree ensemble methods.
Such observation coincides with previous studies \cite{kim-rush-2016-sequence,sun-etal-2019-patient,clark-etal-2019-bam}.

\subsection{Visualization on Heterogeneous Structure}

\begin{figure}[!t]
\centering \includegraphics[width=0.96\columnwidth]{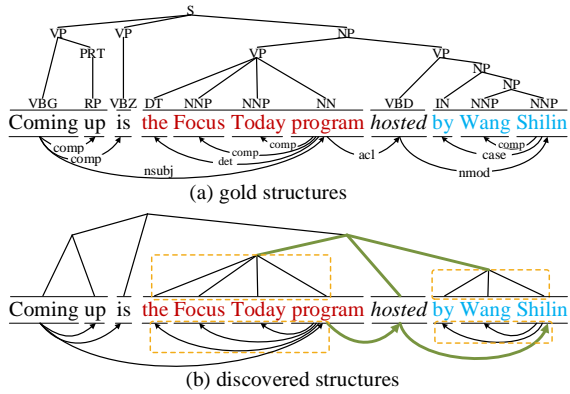}
\caption{
A SRL case where
{\em hosted} is predicate, \textcolor{redi}{the Focus Today program} is \emph{A0}, \textcolor{bluei}{by Wang Shilin} is \emph{A1}.
Bold \textcolor{greeni}{\textbf{green}} lines indicates the edges with higher scores.
}
\label{Visualization1}
\end{figure}

The enhanced structure injection objectives (Eq. (\ref{syn-B-dep}) and (\ref{syn-B-const})) enables the student LSTM to unsupervisedly induce tree structures at the test stage.
To understand how the distilled model promote the mutual learning of heterogeneous structures, we empirically visualize the induced trees based on a test example of SRL.
As shown in Figure \ref{Visualization1}, the discovered dependency structures accurately match the gold tree, and the constituents are highly correlated with the gold one.
Besides, the edges that indicate the two elements are augmented by the learning of each other, which in return enhance the recognition of the spans of elements (yellow dotted boxes), respectively.
For example, the constituent and dependent paths (green lines) linking two minimal target spans,
{\em the Focus Today program} and {\em by Wang Shilin}, are enhanced and echoed with each other, via the core predicate.
This reveals that our method can offer a deeper latent interaction between heterogeneous tree structures.

\section{Conclusion}

We investigated knowledge distillation on heterogeneous tree structures integration for facilitating the NLP tasks, distilling syntactic knowledge into a sequential input encoder, in both output and feature level distillations.
Results on four representative syntax-dependent tasks showed that the distilled student outperformed all standalone tree models, as well as the commonly used ensemble methods, indicating the effectiveness of the proposed method.
Further analysis demonstrated that our method enjoys high robustness and efficiency.

\section*{Acknowledgments}

This work is supported by the National Natural Science Foundation of China (No. 61772378, No. 61702121), 
the National Key Research and Development Program of China (No. 2017YFC1200500), 
the Research Foundation of Ministry of Education of China (No. 18JZD015), 
the Major Projects of the National Social Science Foundation of China (No. 11\&ZD189),
the Key Project of State Language Commission of China (No. ZDI135-112) 
and Guangdong Basic and Applied Basic Research Foundation of China (No. 2020A151501705).

\bibliography{emnlp2020}

\bibliographystyle{acl_natbib}

\end{document}